\documentclass[10pt,twocolumn,letterpaper]{article}

\usepackage[pagenumbers]{cvpr} 
\usepackage{multirow}
\usepackage{array}
\usepackage{tabularx}
\usepackage{makecell}

\definecolor{cvprblue}{rgb}{0.21,0.49,0.74}
\usepackage[pagebackref,breaklinks,colorlinks,allcolors=cvprblue]{hyperref}

\def\paperID{} 
\def\confName{CVPR}
\def\confYear{2026}

\title{FarmMind: Reasoning-Query-Driven Dynamic Segmentation for Farmland Remote Sensing Images}

\author{
\vspace{0.5cm}
Haiyang Wu\textsuperscript{1}, Weiliang Mu\textsuperscript{1}, Jipeng Zhang\textsuperscript{1}, Zhong Dandan\textsuperscript{1}, zhuofei Du\textsuperscript{1}, Haifeng Li\textsuperscript{1}, Tao Chao\textsuperscript{1,\textasteriskcentered}\\
\textsuperscript{1} School of Geosciences and Info-Physics, Central South University, Changsha, China\\
\textsuperscript{\textasteriskcentered} Corresponding author
\\
{\tt\small email: ocean-W@csu.edu.cn}
}

\begin{document}
\maketitle

\begin{abstract}
Existing methods for farmland remote sensing image (FRSI) segmentation generally follow a static segmentation paradigm, where analysis relies solely on the limited information contained within a single input patch. Consequently, their reasoning capability is limited when dealing with complex scenes characterized by ambiguity and visual uncertainty. In contrast, human experts, when interpreting remote sensing images in such ambiguous cases, tend to actively query auxiliary images (such as higher-resolution, larger-scale, or temporally adjacent data) to conduct cross-verification and achieve more comprehensive reasoning. Inspired by this, we propose a reasoning-query-driven dynamic segmentation framework for FRSIs, named FarmMind. This framework breaks through the limitations of the static segmentation paradigm by introducing a reasoning-query mechanism, which dynamically and on-demand queries external auxiliary images to compensate for the insufficient information in a single input image. Unlike direct queries, this mechanism simulates the thinking process of human experts when faced with segmentation ambiguity: it first analyzes the root causes of segmentation ambiguities through reasoning, and then determines what type of auxiliary image needs to be queried based on this analysis. Extensive experiments demonstrate that FarmMind achieves superior segmentation performance and stronger generalization ability compared with existing methods. The source code and dataset used in this work are publicly available at:
https://github.com/WithoutOcean/FarmMind. 
\end{abstract}

\section{Introduction}
\label{sec:intro}
\setlength{\textfloatsep}{5pt}  
\begin{figure}[t]
  \centering
  \includegraphics[width=1\linewidth]{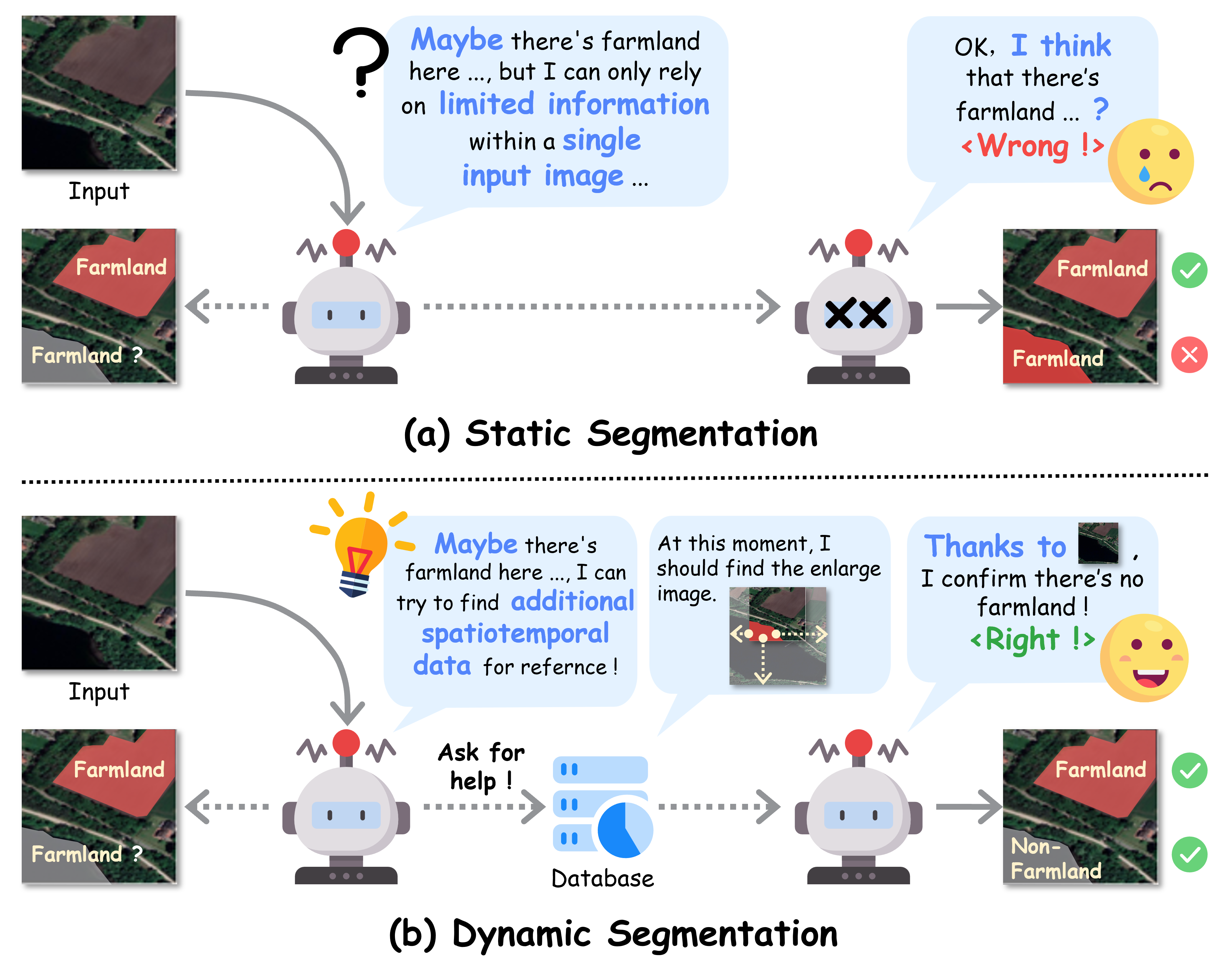}
   \captionsetup{skip=2pt} 
   \caption{Different farmland remote sensing image segmentation paradigms: (a) Static Segmentation; (b) Dynamic Segmentation (Our)}
   \label{fig:fig1}
\end{figure}

Monitoring of farmland resources is a crucial cornerstone for ensuring food security~\cite{47}. Farmland remote sensing image (FRSI) segmentation, as a core technology, provides critical support for the protection of farmland resources and food security~\cite{46}.

For a long time, the field of FRSI segmentation has mainly relied on label-driven segmentation models~\cite{36,37} represented by fully convolutional networks (FCNs) ~\cite{1}, U-Net~\cite{2}, and DeepLab~\cite{3}. However, these methods struggle to deeply understand the intricate relationships between farmland and its surrounding environment, limiting the accuracy of farmland segmentation~\cite{23}. Recently, the rise of language-driven segmentation methods has provided new insights~\cite{4,5,6,40}. Models such as VistaLLM~\cite{7}, LISA~\cite{8}, and EVF-SAM~\cite{9}, through “vision-language” alignment pre-training, transfer the rich semantic priors contained in language to the image segmentation task~\cite{12}. This not only strengthens the model's understanding of the intrinsic properties of farmland, but also enables it to explicitly model the spatial and semantic relationships between farmland and its surrounding environment, thereby improving its generalization capability in complex scenarios.

However, both label-driven and language-driven existing methods generally follow a static segmentation paradigm, where the model's reasoning process rely entirely on the limited information within the current input patch~\cite{21,31}. This results in limited reasoning capabilities when dealing with complex scenes containing ambiguous information. For example, spatially, image cropping may damage the structural integrity of ground objects at boundaries, while the limited contextual relationships within a single patch are insufficient to resolve semantic ambiguities, leading to segmentation errors. Temporally, the visual features of farmland during a specific phenological period are highly similar to those of bare or fallow land. However, a single temporal image patch cannot capture the dynamic evolution of the land, making it insufficient for distinguishing between these categories.

In contrast, when confronted with complex scenes involving segmentation ambiguities, human experts first analyze the root causes of such ambiguities and then retrieve auxiliary images (such as higher-resolution, larger-scale, or temporally adjacent data) to conduct comprehensive and thorough reasoning through cross-validation. Inspired by this, a dynamic segmentation paradigm for farmland is proposed in this study. As shown in Figure~\ref{fig:fig1}, unlike existing static segmentation methods, this paradigm can emulate human experts when faced with interpretation ambiguities: it first conducts reasoning to analyze the root cause of segmentation ambiguity, then proactively initiates queries for relevant auxiliary images based on the analysis, and performs spatio-temporal collaborative reasoning by integrating the auxiliary images. Through the closed-loop process of "reasoning-query-reasoning", it achieves high-precision segmentation.

Furthermore, building upon the theoretical foundation of the dynamic segmentation paradigm, we propose a reasoning-query-driven dynamic framework for farmland segmentation, called FarmMind. By introducing a querying-reasoning mechanism, FarmMind overcomes the limitations of traditional methods that rely solely on limited information within a single input patch for static segmentation. Unlike direct queries, this mechanism simulates the cognitive process of human experts when facing interpretative ambiguities: it first reasons to analyze the root causes of segmentation ambiguities, and then determines what type of auxiliary image needs to be queried based on the analysis. To implement this mechanism, this study constructs an integrated dynamic segmentation pipeline via a multimodal large model (MLLM), comprising basic perception, reasoning queries, and collaborative segmentation. In addition, to support dynamic querying, we have built a global FRSI database, covering multiple countries, which enables efficient retrieval based on geographic coordinates and seasons. The main contributions of this paper are as follows:

\begin{itemize}
    \item We define a dynamic segmentation paradigm. This paradigm breaks through the limitations of existing static segmentation by introducing a querying-reasoning mechanism. It enables models to dynamically and on-demand query external auxiliary images when encountering segmentation ambiguities, thereby compensating for the informational inadequacy of a single input image.
\end{itemize}
\begin{itemize}
    \item We propose a dynamic segmentation framework, FarmMind. Based on a MLLM, this framework establishes a unified dynamic segmentation architecture that integrates basic perception, reasoning queries, and collaborative segmentation, significantly enhancing the robustness of farmland segmentation in complex scenarios.
\end{itemize}
\begin{itemize}
    \item We construct a large-scale FRSI database. To support the training and validation of dynamic segmentation tasks, this study collected and integrated multi-temporal FRSIs covering extensive geographical regions including China, the United States, Germany, and other countries, establishing a large-scale FRSI database.
\end{itemize}

\section{Related work}
\label{sec:formatting}

\subsection{Label-driven FRSI segmentation}

Label-driven FRSI segmentation methods establish a static mapping between image features and predefined labels~\cite{14}, performing well in conditions where farmland is distinct and scene is regular. These models often incorporate attention mechanisms~\cite{16}, multi-scale feature representations~\cite{21}, and boundary enhancement strategies~\cite{19,20} to improve performance. However, they remain limited to modeling local visual features and fail to capture the complex semantic and spatial relationships between farmland and its surrounding environment. Moreover, such methods follow a static interpretation paradigm, relying on fixed image patches for decision-making, which limits their adaptability to dynamic scene changes.

\subsection{Language-driven FRSI segmentation}

In FRSI segmentation, language-driven multimodal large language models (MLLMs) have significantly enhanced reasoning and generalization capabilities in complex scenarios by incorporating the semantic comprehension of language~\cite{10,11,22,38,39}. As a carrier of knowledge, language possesses strong semantic associations and logical expressiveness~\cite{43,44,45}, enabling abstract representation and compositional reasoning of farmland’s spatiotemporal characteristics from multiple dimensions such as seasonal variation, topographic features, and surrounding environments. Based on this idea,  FSVLM~\cite{23} employs a MLLM (LLaVA~\cite{32}) as the reasoning core to guide SAM~\cite{24} in farmland segmentation, demonstrating superior performance over traditional methods in complex agricultural scenes. However, existing MLLM-based methods remain confined to a static interpretative framework, whereas real farmland environments exhibit significant spatiotemporal dynamics, continuously influenced by crop growth, climate, and human activities. This fundamental contradiction between the static mechanism and the dynamic environment limits the model’s ability to handle complex situations such as boundary ambiguity and class confusion.

\subsection{Image retrieval-augmented generation}

In recent years, to enhance the generalization capability of models in open scenarios, some studies~\cite{48, 49, 50} have begun to introduce Retrieval-Augmented Generation (RAG) technology into the field of image processing. For example, ImageRAG~\cite{51} integrates the RAG mechanism with diffusion models to alleviate performance bottlenecks when generating rare or unseen concepts. A similar approach~\cite{52} has also been applied to address challenges in visual question answering tasks for ultrahigh resolution remote sensing images. Although existing studies have verified that image RAG aids model understanding, two shortcomings remain: first, there is a lack of explicit modeling regarding “why retrieval is needed” and “how to collaboratively utilize retrieval results for deep reasoning”; second, in tasks highly dependent on spatiotemporal context and semantic consistency, such as FRSI segmentation, an active retrieval mechanism based on the characteristics of farmland has not yet been established. Therefore, there is an urgent need to develop a dynamic framework capable of proactively diagnosing uncertainty, triggering precise retrieval on demand, and achieving cross-image collaborative reasoning. The FarmMind proposed in this paper is an in-depth exploration aimed at addressing the aforementioned issues. Unlike existing methods that directly retrieve queries, FarmMind, by incorporating the characteristics of farmland, can deeply analyze the causes of segmentation ambiguity when faced with it and initiate on-demand queries accordingly. This upgrades retrieval augmentation from a passive auxiliary tool to an active cognitive means, promoting the transition of FRSI segmentation from static to dynamic.

\section{Methodological}
\label{sec:method}

\subsection{Overview}

\begin{figure*}[!htb]
  \centering
  \includegraphics[width=0.85\linewidth]{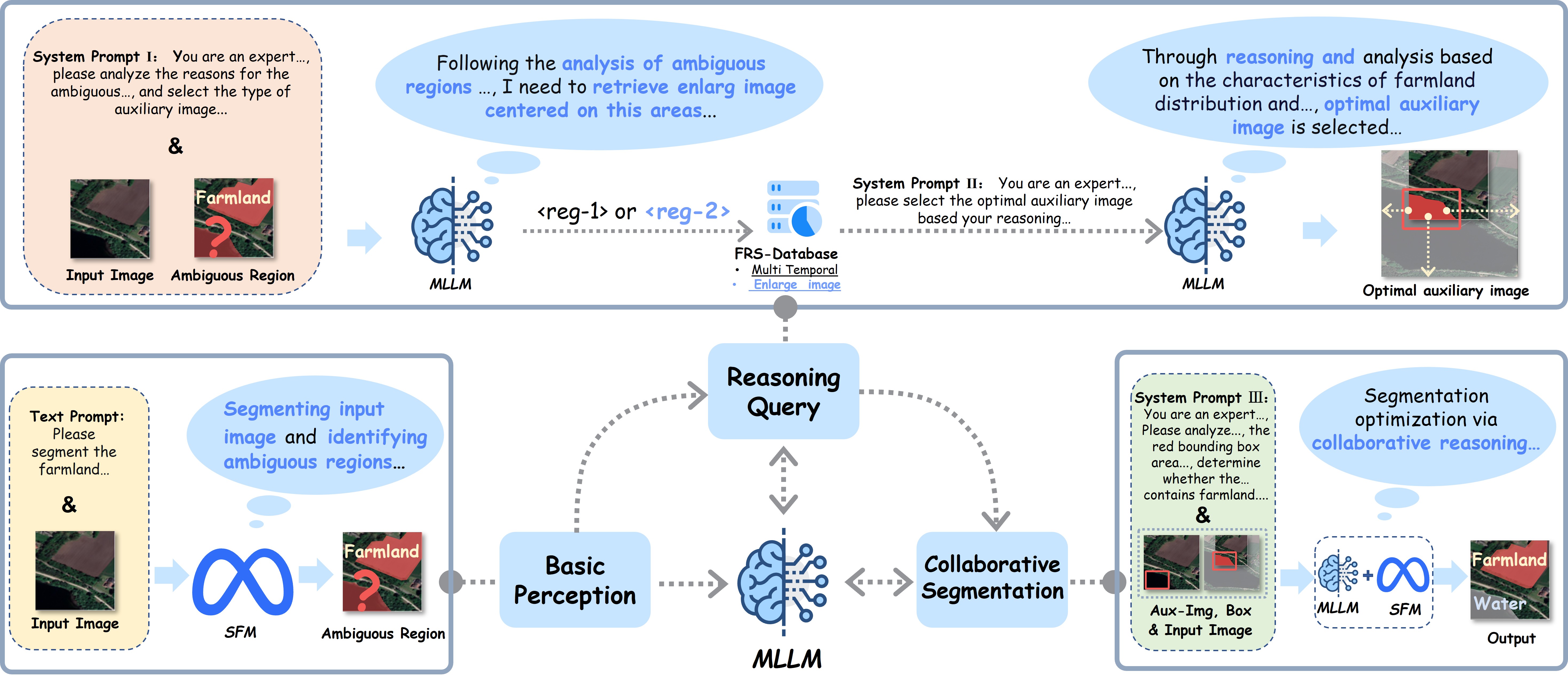}
  \captionsetup{skip=2pt} 
   \caption{Overview of dynamic segmentation for farmland remote sensing images. (FSM stands for foundation segmentation model; MLLM stands for multimodal large language model )}
   \label{fig:Fig2}
\end{figure*}
To overcome the limitations of the static segmentation paradigm, this paper proposes a dynamic segmentation framework, FarmMind. As shown in Figure~\ref{fig:Fig2}, FarmMind is a dynamic segmentation framework consisting of three stages: basic perception, reasoning query, and collaborative segmentation. In the basic perception stage, FarmMind performs an initial segmentation of the remote sensing image and identifies areas with high uncertainty and ambiguity. In the reasoning query stage, attribution reasoning analysis is conducted for these ambiguous areas, and relevant auxiliary image sets are retrieved as needed, automatically selecting the optimal auxiliary images. In the collaborative segmentation stage, the system integrates the selected auxiliary images for collaborative reasoning, accurately determining and refining the segmentation of the ambiguous regions, ultimately outputting the optimized farmland segmentation results. A complete workflow example of FarmMind is shown in Figure~\ref{fig:Fig3}. A more detailed and complete example can be found in \textbf{Supplementary Material A}.

\begin{figure*}[t]
  \centering
  \includegraphics[width=0.6\linewidth]{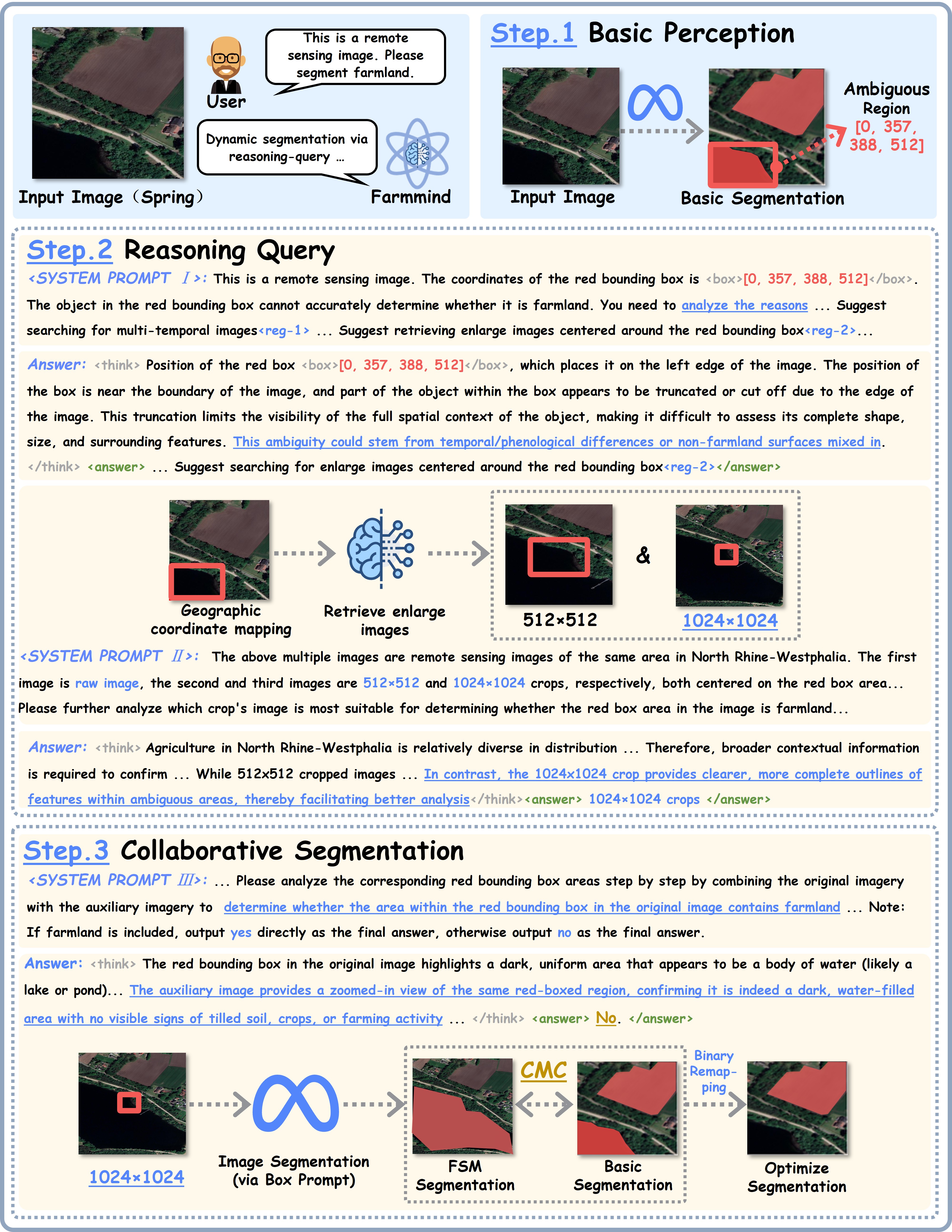}
  \captionsetup{skip=2pt} 
   \caption{The complete workflow example of FarmMind. (Conditional mask correction (CMC) strategy and binary remapping are detailed in Section~\ref{subsec:3.4})}
   \label{fig:Fig3}
\end{figure*}

\subsection{Basic perception}
The basic perception aims to obtain the initial segmentation result of the farmland remote sensing image (FRSI) and the corresponding segmentation ambiguous regions. It mainly comprises two steps: basic segmentation and ambiguous region selection.

\textbf{Basic segmentation.} We first perform efficient fine-tuning of foundation segmentation model (FSM) on the farmland image-text dataset FarmSeg-VL~\cite{26}. The fine-tuned model is then employed to generate the basic segmentation mask for FRSI, along with the corresponding confidence map.

\textbf{Ambiguity region selection.}  To identify potential ambiguous regions, we propose a confidence-threshold-based ambiguity region selection strategy. Specifically, given the confidence scores from the base segmentation, we define a confidence threshold interval\(\left[-T, T\right]\). Pixels with confidence values within this interval are assigned a value of 1, while others are set to 0, thereby generating a binary confidence mask. Assuming that the confidence score output by FSM is \(C(x, y)\), where \(x\) and \(y\) denote the horizontal and vertical coordinates of the pixels in the confidence mask, respectively. The binarized low-confidence mask \(M(x, y)\) can be expressed as:
\[
M(x, y) = 
\begin{cases}
1, & \text{if } C(x, y) \in [-T, T], \\
0, & \text{otherwise.}
\tag{1}
\end{cases}
\]

Subsequently, all connected regions with a label value equal to 1 in the mask are extracted. For each connected region \( R_i \) (where \( i \) denotes the index of the connected region), the minimum bounding box is calculated. To exclude noise and interference from irrelevant small regions, only regions with the bounding boxes area within the preset range \(\left[ S, S + s \right]\), where $S$ is the minimum area threshold and $s$ is the allowed area increment. 

\subsection{Reasoning query}
For the segmentation regions already identified as ambiguous, the query reasoning stage aims to eliminate uncertainty by answering three core questions: when to query, what to query, and how to query. It mainly consists of three key steps: attribution reasoning, data query, and optimal auxiliary image selection.

\textbf{Attribution reasoning.} To enable FarmMind to autonomously reason and analyze the causes of segmentation ambiguities, similar to human reasoning, and thereby trigger proactive queries for auxiliary images, we use multimodal large language model (MLLM) as the reasoning-query model and introduce a hierarchical system prompt \MakeUppercase{\romannumeral 1\relax} (see \textbf{Supplementary Material B}). This prompt is designed to guide MLLM in deeply analyzing the underlying causes of segmentation ambiguities and, based on the analysis results, proactively initiate queries for auxiliary images. To ensure that MLLM can accurately identify and analyze ambiguity regions, we not only add red bounding boxes around the ambiguous areas in the input image but also provide the specific coordinates of the target boxes in the system prompt \MakeUppercase{\romannumeral 1\relax}. Additionally, to effectively guide MLLM in cause analysis, we explicitly define two major reasoning directions in system prompt \MakeUppercase{\romannumeral 1\relax}: first, visual ambiguity caused by spatiotemporal heterogeneity, manifested as unstable visual features due to crop phenology or land cover changes; second, incomplete semantic information in boundary regions, where objects located at image edges or near cropped boundaries suffer from missing contextual information, thereby affecting class discrimination. Finally, to standardize the subsequent query process, we specify the data types returned by the model using \texttt{\textless reg-1\textgreater} and \texttt{\textless reg-2\textgreater} tags.

\textbf{Data query.} During the data query phase, explicit data query instructions are generated based on the attribution results from MLLM, and relevant auxiliary images are extracted from the FRSI database (See section~\ref{subsec:3.5}) constructed in this study according to these instructions. Specifically, the identifiers \texttt{\textless reg-1\textgreater} and \texttt{\textless reg-2\textgreater} are used in the query instructions to denote the retrieval of other temporal images or enlarge images, respectively. The query command must precisely specify parameters such as the geographic coordinate range and the target image type to efficiently retrieve from external databases a set of auxiliary images highly relevant to the current task. It should be noted that the enlarge image queried in this work refers to a larger-scale FRSI centered on the ambiguous region. Upon receiving the structured query command, FarmMind automatically parses the geographic coordinates and the data type specified in the command to perform precise matched searches and crops the required auxiliary images from the FRSI database constructed in this study. It is worth noting that the data query phase in this research is not limited to the FRSI database alone. Instead, it can flexibly initiate cross-database and multi-source data queries based on the geographic coordinates of the input data and the type of data retrieval. Therefore, this mechanism can be regarded as an open-world query tool, capable of dynamically responding to various geographic information needs, significantly enhancing FarmMind's adaptability and scalability.  

\textbf{Optimal auxiliary image selection.} To avoid redundant computations and maximize information gain, after obtaining the candidate auxiliary images, this study utilizes system prompt \MakeUppercase{\romannumeral 2\relax} (see \textbf{Supplementary Material B}) to guide MLLM in reasoning to select the optimal auxiliary image. Specifically, when multiple candidate auxiliary images are available, FarmMind initiates a selection decision process. This system prompt \MakeUppercase{\romannumeral 2\relax} guides MLLM to perform comprehensive reasoning on the candidate images, considering multiple features such as temporal continuity, spatial alignment, and semantic complementarity. The optimal auxiliary image is then selected, accompanied by the corresponding rationale, ensuring that the information gain from the introduced auxiliary information is maximized.

\subsection{Collaborative segmentation}
\label{subsec:3.4}
The core issue that collaborative segmentation aims to address is how to fully exploit and effectively utilize the complementary information from external auxiliary images to optimize the segmentation results. This process mainly includes two key steps: multi-image reasoning and dynamic correction.

\textbf{Multi-image reasoning.}  In this stage, guided by System Prompt \MakeUppercase{\romannumeral 3\relax} (see \textbf{Supplementary Material B}), MLLM conducts a joint analysis of the original image \( x_{\text{img}} \) and the optimal auxiliary image \( \widetilde{x}_{\text{img}} \). Rather than simply overlaying pixels, it deeply explores the spatiotemporal and semantic relationships exhibited by the two images within the ambiguous regions. For instance, when the segmentation ambiguity arises from phenological variations, an auxiliary image from other temporal phases can provide continuity evidence of crop growth. By integrating these complementary pieces of information, MLLM ultimately generates a definitive text description \( y_{\text{txt}} \) , indicating whether the ambiguous region contains farmland.

\textbf{Dynamic correction.} This process first employs FSM to segment the optimal auxiliary image \( \widetilde{x}_{\text{img}} \), obtaining a refined segmentation of the ambiguous region. Then, through a conditional mask correction (CMC) mechanism, it dynamically rectifies the basic segmentation mask $\hat{y}$ (from the basic perception stage) by combining the textual output  \( y_{\text{txt}} \) from multi-image collaborative reasoning with the segmentation result of FSM.
Specifically, the auxiliary image \( \widetilde{x}_{\text{img}} \) and the bounding box coordinates of its corresponding ambiguous region are first fed into FSM to generate a fine-grained binary mask \( y_s \). Then, the CMC mechanism corrects \( \hat{y} \) according to the text description \( y_{\text{txt}} \): If \( y_{\text{txt}} \) is “yes", indicating that the region contains farmland, mask addition is performed; otherwise, a mask subtraction operation is applied. This process can be formulated as:
\[
y_c = \begin{cases} 
\hat{y} + y_s & \text{if } y_{\text{txt}} = \text{"yes"} \\
\hat{y} - y_s & \text{if } y_{\text{txt}} = \text{"no"}
\end{cases} \tag{2}
\]

where \( y_c \) denotes the intermediate corrected mask. Since the operations may cause pixel values to exceed the binary range, a binary remapping is performed to obtain the final refined mask \( y_f \). For each pixel location \( (i, j) \):

\[
y_{f(i,j)} = \min \left( \max \left( 0, y_{c(i,j)} \right), 1 \right) \tag{3}
\]

The entire process fully leverages the spatiotemporal complementarity between the original image and the auxiliary image, achieving stable and interpretable optimization of the segmentation results.

\subsection{A FRSI database collection and organization}
\label{subsec:3.5}
\begin{figure*}[t]
  \centering
  \includegraphics[width=0.8\linewidth]{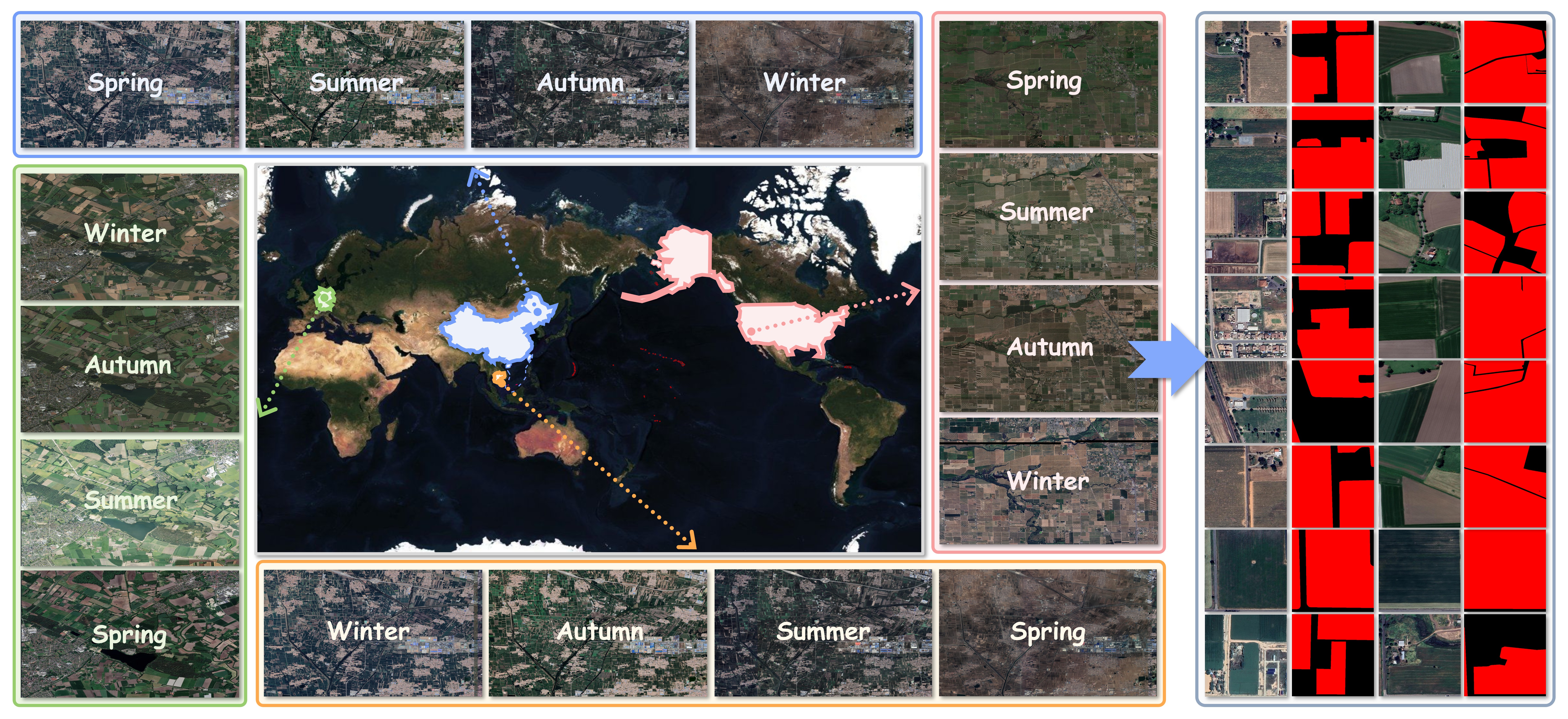}
  \captionsetup{skip=2pt} 
   \caption{Examples of partial data from the FRSI database}
   \label{fig:Fig4}
\end{figure*}

To support the dynamic segmentation task for FRSI, this study constructs a large-scale and dynamically retrievable FRSI database (More details see \textbf{Supplementary Material C}). This database integrates 112 scenes of wide-swath remote sensing images, with varying image sizes, and pixel counts ranging from tens of millions to hundreds of millions. It covers major agricultural regions in China, as well as representative areas in the United States, Cambodia, and Germany. The images span multiple climate zones and farming practices, with spatial resolutions ranging from 0.5 meters to 2 meters, and exhibit a broad temporal span with strong seasonal representativeness. Furthermore, as shown in Figure \ref{fig:Fig4}, to facilitate testing and validation, one seasonal image from each region in the database is selected and cropped into 512×512 image patches, each accompanied by a set of high-precision ground-truth mask annotations, resulting in a total of 11,131 samples. The original uncut panoramic images retain complete geographic coordinate information, serving as the query objects. To enhance the efficiency of image retrieval, this paper designs a hierarchical indexing strategy: the first level divides images by data type (multi-temporal images, enlarge images), and the second level organizes them by administrative. During retrieval, the system first filters candidate images based on data type, then refines the selection by province, and finally locates the specific image based on the target geographic coordinates, cropping and outputting the required result. It is worth noting that the patch annotation data for the Chinese region in this database is  sourced from FarmSeg-VL~\cite{26}, and we only collected wide-swath images of the corresponding areas across different time phases.

\section{Experiments and results}
\label{sec:exp}

\subsection{Experimental setup}

\textbf{Implementation details.} All experiments are conducted on two NVIDIA A6000 GPUs. Unless otherwise specified, all models are trained on the FarmSeg-VL~\cite{26} and directly evaluated using the test set of this dataset (i.e., the Chinese region covered in our constructed FRSI database). The query–reasoning model (QRM) in this work is implemented by directly calling the Qwen-VL-Max API. In the basic segmentation stage, the confidence threshold was configured to \([-1, 1]\), and the area range for ambiguous regions was configured to \([5000, 100{,}000]\). And we adopt FSVLM as the foundation segmentation model (FSM), and all its hyperparameters follow the original paper. In the collaborative segmentation stage, we directly employ SAM2~\cite{13} as the FSM, with the model parameters set to  \texttt{sam2.1\_hiera\_large}. It is worth noting that, within the proposed framework, both the QRM and the FSM are flexible and can be replaced with other functionally equivalent alternatives.

\textbf{Evaluation metrics.} Following existing studies on farmland segmentation, this paper adopts mean accuracy (mACC), mean intersection over union (mIoU), F1-score, and Recall as evaluation metrics.





\subsection{Method comparison}
To comprehensively evaluate the performance of the proposed method in the farmland remote sensing image (FRSI) segmentation task, this paper compares label-driven segmentation methods (U-Net~\cite{2}, DeepLabv3+~\cite{27}, SegFormer~\cite{28}), language-based segmentation methods (PixelLM~\cite{29}, LaSagnA~\cite{30}, FSVLM~\cite{23}).

\begin{table*}[t]
\captionsetup{skip=2pt} 
\centering
\caption{Comparison of segmentation performance among different methods.}
\label{tab:tab1}
\renewcommand{\arraystretch}{1}
\setlength{\tabcolsep}{7pt}
\resizebox{0.78\textwidth}{!}{
\begin{tabular}{lllcccc}
\hline
\multicolumn{3}{c}{\textbf{Methods}} & \textbf{mAcc(\%)} & \textbf{mIoU(\%)} & \textbf{F1(\%)} & \textbf{Recall(\%)} \\
\hline
\multirow{6}{*}{\parbox[c]{28mm}{\raggedright\textit{Static segmentation}}}
  &                       & U\mbox{-}Net      & 75.18 & 59.91 & 74.84 & 75.18 \\
  & Label-driven          & DeepLabv3+        & 79.94 & 66.23 & 79.61 & 79.94 \\
  &                       & SegFormer         & 86.45 & 76.13 & 86.44 & 86.45 \\
\cline{3-7}
  &                       & PixelLM           & 86.38 & 75.88 & 86.27 & 87.39 \\
  & Language-driven            & LaSagnA           & 89.70 & 81.35 & 89.71 & 89.08 \\
  &                       & FSVLM             & 90.02 & 83.39 & 90.12 & 90.51 \\
\hline
\multicolumn{1}{l}{\raggedright\textit{Dynamic segmentation}}
  &                       & FarmMind  & \textbf{92.85} & \textbf{84.43} & \textbf{91.25} & \textbf{93.04} \\
\hline
\end{tabular}
}
\end{table*}

\begin{figure*}[t]
  \centering
  \includegraphics[width=0.78\linewidth]{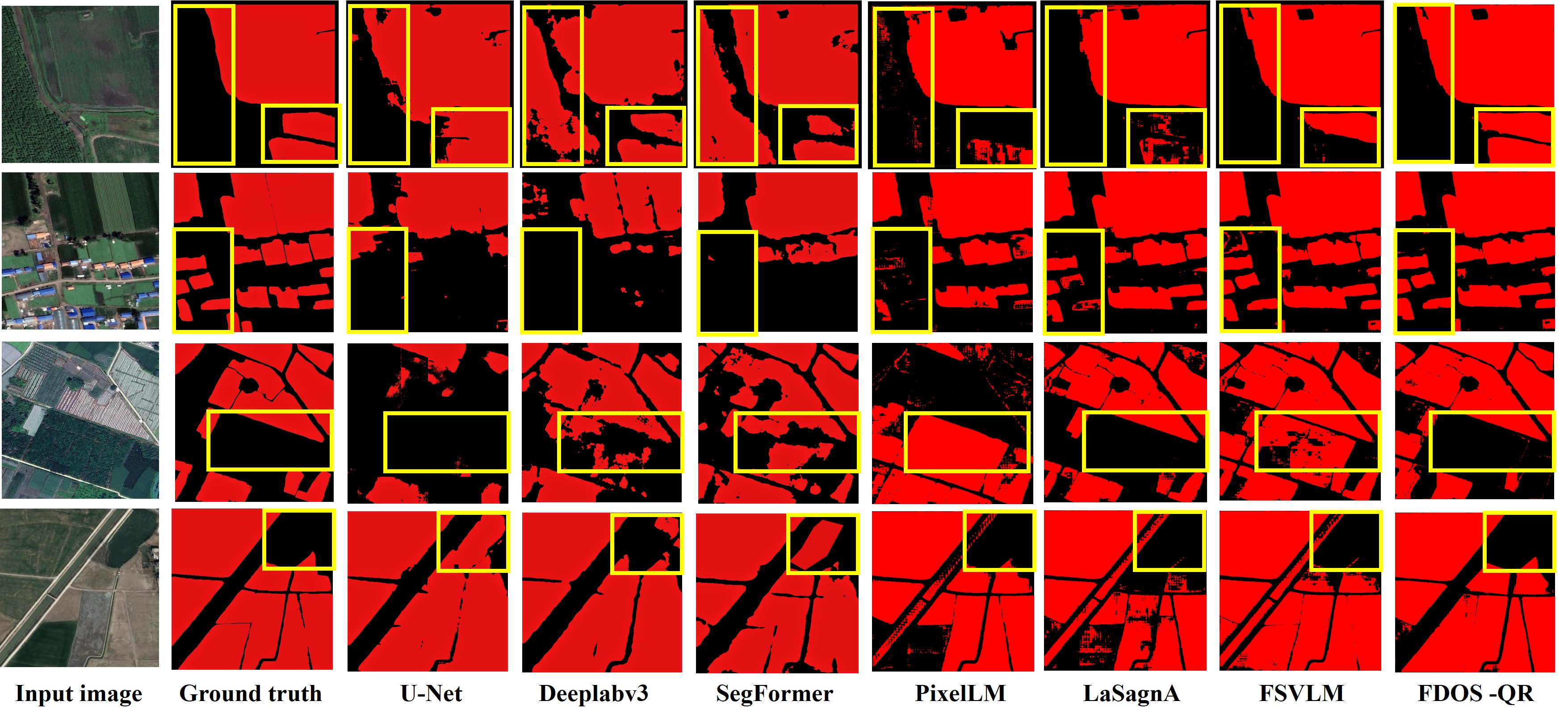}
    \captionsetup{skip=2pt} 
   \caption{Segmentation maps of different methods}
   \label{fig:Fig5}
\end{figure*}

As shown in Table \ref{tab:tab1}, the overall mixed-region segmentation results on the FarmSeg-VL test set (detailed results for each region are provided in \textbf{Supplementary Material D}) indicate that label-driven segmentation methods perform poorly in complex agricultural landscapes, with U-Net achieving an mIoU of only 59.91\%. Although DeepLabv3 and SegFormer improve performance through multi-scale feature fusion, they still lack the ability to capture the intrinsic semantics of farmland due to their reliance on single-modal visual cues. In contrast, language-driven segmentation methods significantly enhance segmentation performance by jointly modeling linguistic and visual information. However, these methods still operate within a static segmentation paradigm, making them less adaptable to the spatiotemporal dynamics of farmland environments. The proposed FarmMind outperforms all existing methods across all evaluation metrics. This remarkable improvement can be attributed to the introduction of the reasoning-query mechanism, which strengthens the model’s understanding of farmland spatiotemporal relationships by retrieving and integrating external auxiliary information. Furthermore, Figure~\ref{fig:Fig5} presents segmentation visualization results for representative scenarios. It can be observed that both label-driven segmentation methods and language-driven methods tend to exhibit boundary adhesion or omission issues in patch-cut regions or areas with land-cover confusion caused by spatiotemporal heterogeneity. In contrast, FarmMind effectively mitigates uncertainty in low-confidence regions by supplementing contextual information through cross-spatiotemporal querying, thereby producing more coherent and accurate farmland segmentation results.

\subsection{Ablation and variant analysis}
\textbf{Ablation study on different FSMs.} As shown in Table \ref{tab:tab2}, a stronger FSM provides better initial segmentation results. When Qwen-VL-Max is introduced as the reasoning-query model(RQM),  the segmentation performance of all FSMs demonstrates significant improvement, demonstrating the general enhancement capability of the proposed  dynamic framework. It is worth noting that although stronger FSMs produce fewer initial errors, the correction effect after adding Qwen-VL-Max shows a downward trend. This suggests that for high-accuracy FSMs, the  dynamic framework encounters more complex error patterns, making correction more challenging. Furthermore, excessive modification may introduce additional false alarms, leading to a reduction in correction accuracy. This observation suggests that the correction strategy should be dynamically adjusted according to the characteristics of different FSMs to avoid unnecessary interference with high-confidence regions.
\begin{table}[t]
\captionsetup{skip=2pt} 
\centering
\caption{Performance comparison of the proposed framework with different foundation segmentation models. (``--'' indicates not applicable; RQM stands for reasoning-query model; FSM stands for foundation segmentation model)}
\label{tab:tab2}
\resizebox{0.47\textwidth}{!}{  
\begin{tabular}{cccccc}
\hline
\textbf{RQM} & \textbf{FSM} & \textbf{mACC(\%)} & \textbf{mIoU(\%)}& \textbf{F1(\%)} & \textbf{Recall(\%)}\\
\hline
--           & PixelLM   & 86.38 & 75.88 & 86.27 & 87.39  \\
Qwen-VL-Max  & PixelLM   & 87.94 & 78.96 & 88.08 & 89.96  \\
\cline{1-6}
--           & LaSagnA   & 89.70 & 81.35 & 89.71 & 89.08  \\
Qwen-VL-Max  & LaSagnA   & 91.57 & 83.09 & 91.48 & 91.01  \\
\cline{1-6}
--           & FSVLM     & 90.02 & 83.39 & 90.12 & 90.51 \\
Qwen-VL-Max  & FSVLM     &\textbf{ 92.85} & \textbf{84.43}&\textbf{ 91.25} & \textbf{93.04}\\
\hline
\end{tabular}
}
\end{table}

\textbf{Ablation study on different RQMs.} As shown in Table \ref{tab:tab3}, the ablation study evaluates different RQM using FSVLM as the FSM. The results demonstrate a consistent performance improvement across all metrics after introducing various RQMs. Notably, Qwen-VL-Max shows the most significant improvements, achieving the highest mACC (92.85\%) and F1 score (91.25\%), along with the best recall (93.04\%). Similarly, Qwen3-VL-30B delivers the highest mIoU (84.93\%), emphasizing its strength in enhancing region overlap. Other models, such as GPT-4o and Qwen2.5-VL-72B, also show notable improvements. In conclusion, the proposed FarmMind effectively enhances the performance of the FSM, and its correction effect is influenced by the capabilities of the RQM.

\begin{table}[t]
\captionsetup{skip=2pt} 
\centering
\caption{Ablation study on different reasoning-query models using FSVLM as the foundation segmentation model. (``--'' indicates not applicable; RQM stands for reasoning-query model; FSM stands for foundation segmentation model)}
\label{tab:tab3}
\resizebox{0.47\textwidth}{!}{  
\begin{tabular}{ccccccc}
\hline
\textbf{RQM} & \textbf{FSM} & \textbf{mACC(\%)} & \textbf{mIoU(\%)}& \textbf{F1(\%)} & \textbf{Recall(\%)}\\
\hline
--           & FSVLM     & 90.02 & 83.39 & 90.12 & 90.51\\
Qwen-VL-Max  & FSVLM     & \textbf{92.85} & 84.43 &\textbf{ 91.25} & \textbf{93.04}\\
GPT-4o~\cite{33}      & FSVLM     & 91.74 & 84.64 & 91.02 & 92.87\\
Qwen2.5-VL-72B~\cite{34} & FSVLM   & 91.79 & 84.72 & 91.53 & 92.64\\
Qwen3-VL-30B~\cite{35}   & FSVLM   & 91.92 & \textbf{84.93} & 91.04 & 92.96\\
\hline
\end{tabular}
}
\end{table}

\begin{figure*}[t]
  \centering
  \includegraphics[width=0.85\linewidth]{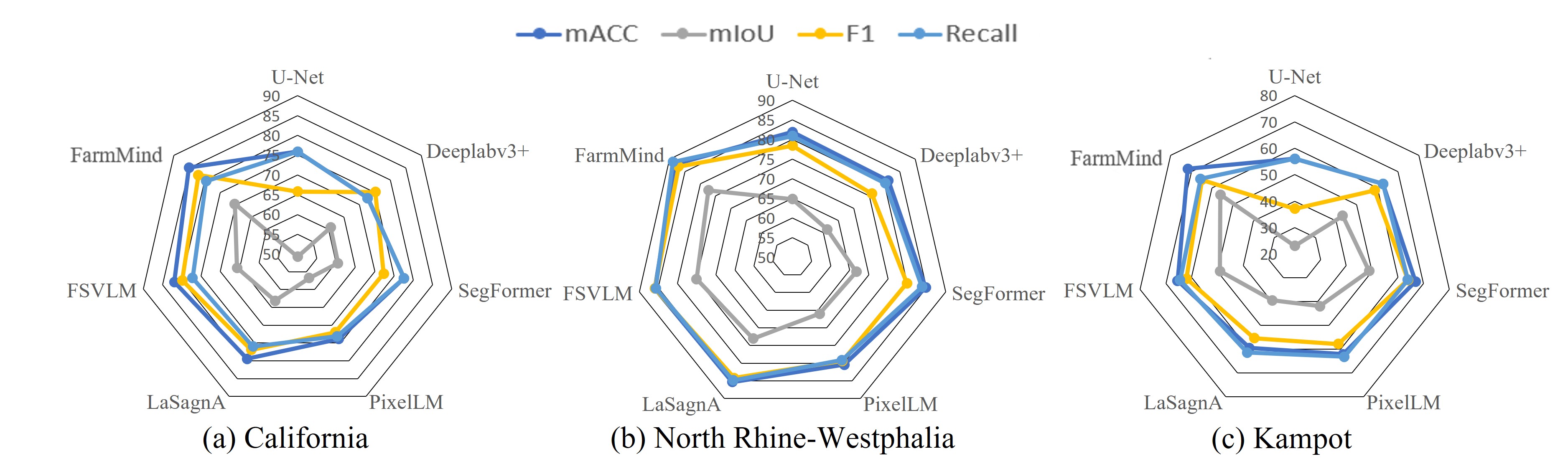}
    \captionsetup{skip=2pt} 
   \caption{Cross domain performance comparison of different methods}
   \label{fig:Fig6}
\end{figure*}

\subsection{Cross-domain generalization}
To comprehensively evaluate the model’s generalization capability in real-world complex scenarios, cross-domain experiments were conducted in three regions that differ significant from the Chinese training areas: California (United States), North Rhine-Westphalia (Germany), and Kampot (Cambodia). As shown in Figure \ref{fig:Fig6}, traditional label-driven methods perform poorly across all metrics, with their radar plots showing a clear inward contraction. This indicates that the closed segmentation paradigm, which relies on static visual priors suffers from severely limited generalization ability when confronted with cross-domain challenges such as farmland fragmentation and land-cover confusion. In contrast, language-driven methods demonstrate superior performance by leveraging the enhanced semantic understanding provided by language priors, which effectively alleviates class ambiguity. The proposed method forms the outermost closed contour on the radar plot across all evaluation metrics, significantly outperforming the comparative methods and thereby demonstrating its superior cross-domain generalization capability.

\section{Conclusion}
\label{sec:clu}
To overcome the limitations of existing methods in farmland remote sensing image (FRSI) segmentation, which rely on static segmentation, we propose a dynamic segmentation paradigm and constructs the corresponding dynamic segmentation framework, FarmMind. This framework introduces a reasoning-query mechanism, enabling the model to proactively query auxiliary images on demand through reasoning when faced with segmentation ambiguities, thereby effectively addressing typical issues such as semantic ambiguity caused by image cutting and ground object confusion due to phenological changes. Experimental results demonstrate that FarmMind consistently outperforms existing methods across representative agricultural regions in multiple countries and exhibits superior cross-domain generalization capability. Overall, this study alleviates the constraints of the static segmentation paradigm and provides a new perspective for developing FRSIs intelligent interpretation framework with spatiotemporal awareness through its demonstrated capabilities in active perception, dynamic querying, and collaborative reasoning.
{
    \small
    \bibliographystyle{ieeenat_fullname}
    \bibliography{main}
}

\clearpage
\setcounter{page}{1}
\maketitlesupplementary
\setcounter{section}{0} 
\renewcommand{\thesection}{\Alph{section}} 
\setcounter{table}{0} 
\renewcommand{\thetable}{\Roman{table}}

\begin{figure*}[htbp]
  \centering
  \includegraphics[width=0.95\linewidth]{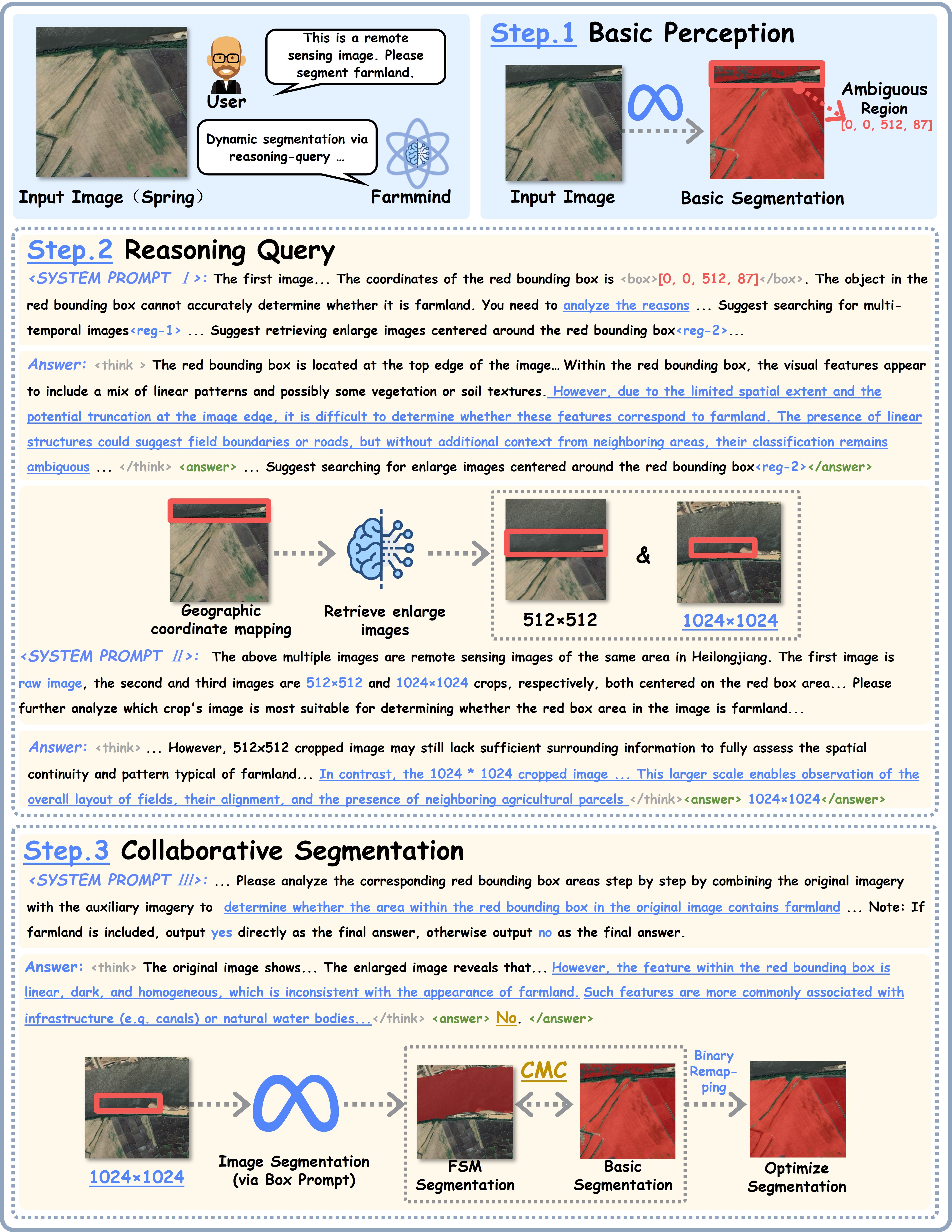}
   \captionsetup{skip=2pt} 
   \caption{The complete workflow example of FarmMind (Retrieve enlarge images).}
   \label{fig:SM-fig1}
\end{figure*}

\begin{figure*}[htbp]
  \centering
  \includegraphics[width=0.95\linewidth]{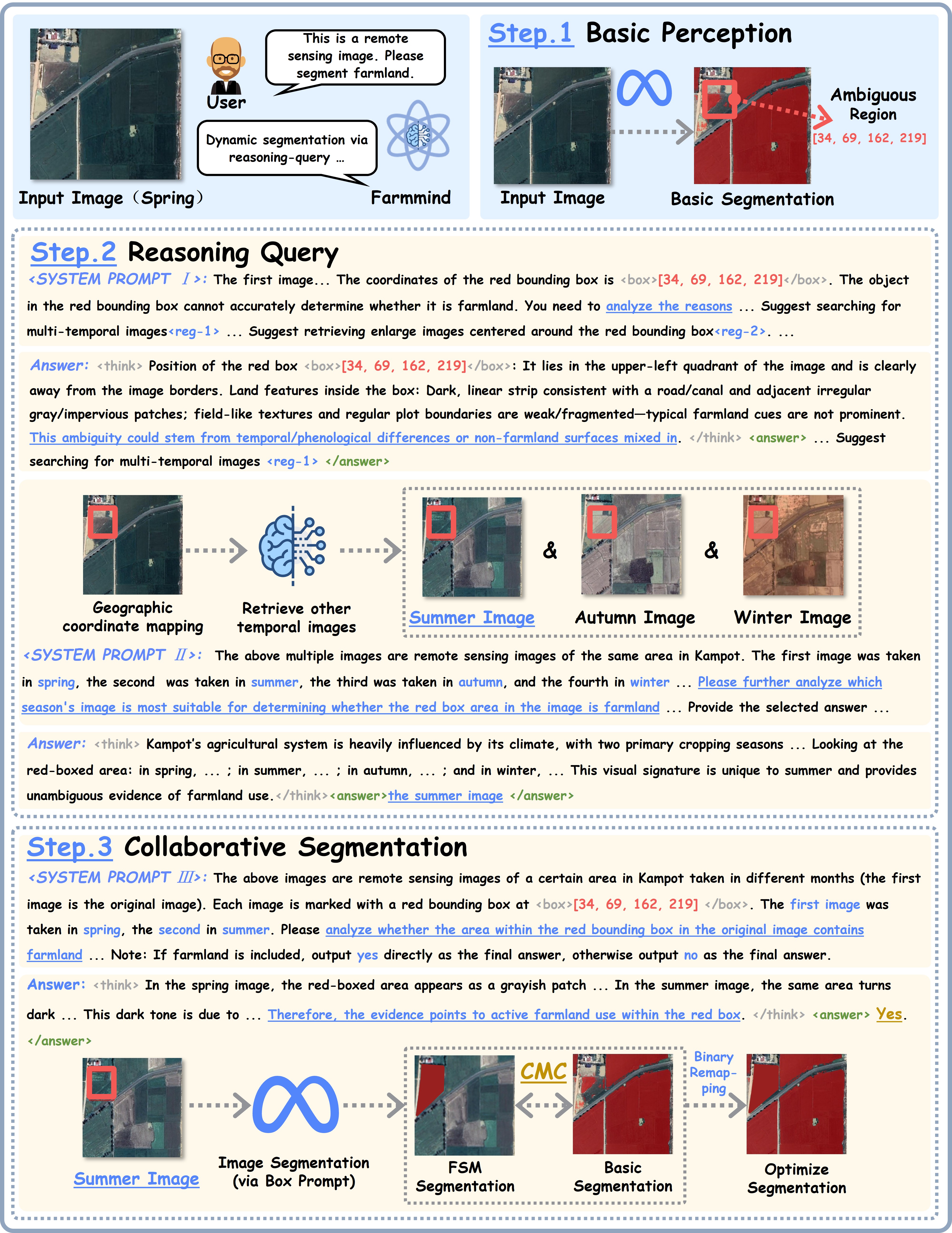}
   \captionsetup{skip=2pt} 
   \caption{The complete workflow example of FarmMind (Retrieve multi-temporal images).}
   \label{fig:SM-fig2}
\end{figure*}
\textbf{In the supplementary material, we provide the following additional content:}

\begin{itemize}
    \item A complete working example of the proposed FarmMind framework, as shown in~\textbf{Section A}.
\end{itemize}

\begin{itemize}
    \item Detailed system prompt examples included in the proposed FarmMind framework, as shown in~\textbf{Section B}.
\end{itemize}

\begin{itemize}
    \item Detailed information on the large-scale farmland remote sensing image database we constructed, as shown in~\textbf{Section C}.
\end{itemize}

\begin{itemize}
    \item Additional experiments and discussions, as shown in~\textbf{Section D}.
\end{itemize}

\section{The complete workflow of FarmMind}
\label{sec:A}
In this section, we provide additional working examples of the FarmMind framework, including how to retrieve enlarge images as auxiliary data (see Figure~\ref{fig:SM-fig1}) and how to retrieve and utilize multi-temporal images as auxiliary data (see Figure~\ref{fig:SM-fig2}). Specifically, when the FarmMind framework identifies ambiguous regions in an image, it analyzes the causes of such ambiguity through a series of reasoning processes and actively queries relevant auxiliary images. Based on the retrieved images, FarmMind further selects the most suitable auxiliary image from the candidate set. Finally, it performs collaborative reasoning by integrating the original image with the auxiliary image, thereby achieving more accurate collaborative reasoning and decision-making.

\section{Hierarchical system prompt}
\label{sec:B}
\begin{figure*}[htbp]
  \centering
  \includegraphics[width=0.8\linewidth]{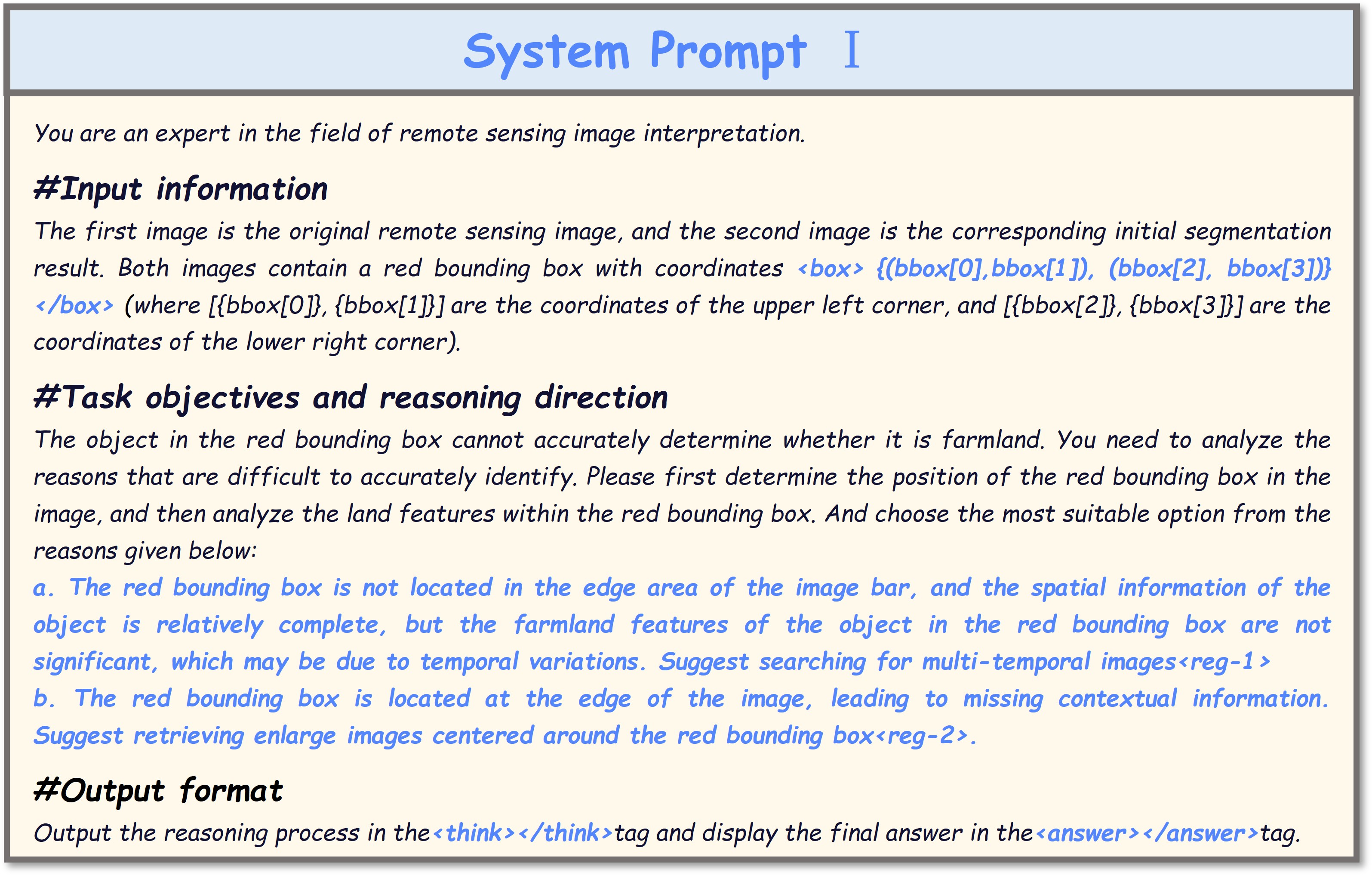}
   \captionsetup{skip=2pt} 
   \caption{System Prompt I.}
   \label{fig:prompt1}
\end{figure*}

\begin{figure*}[htbp]
  \centering
  \includegraphics[width=0.8\linewidth]{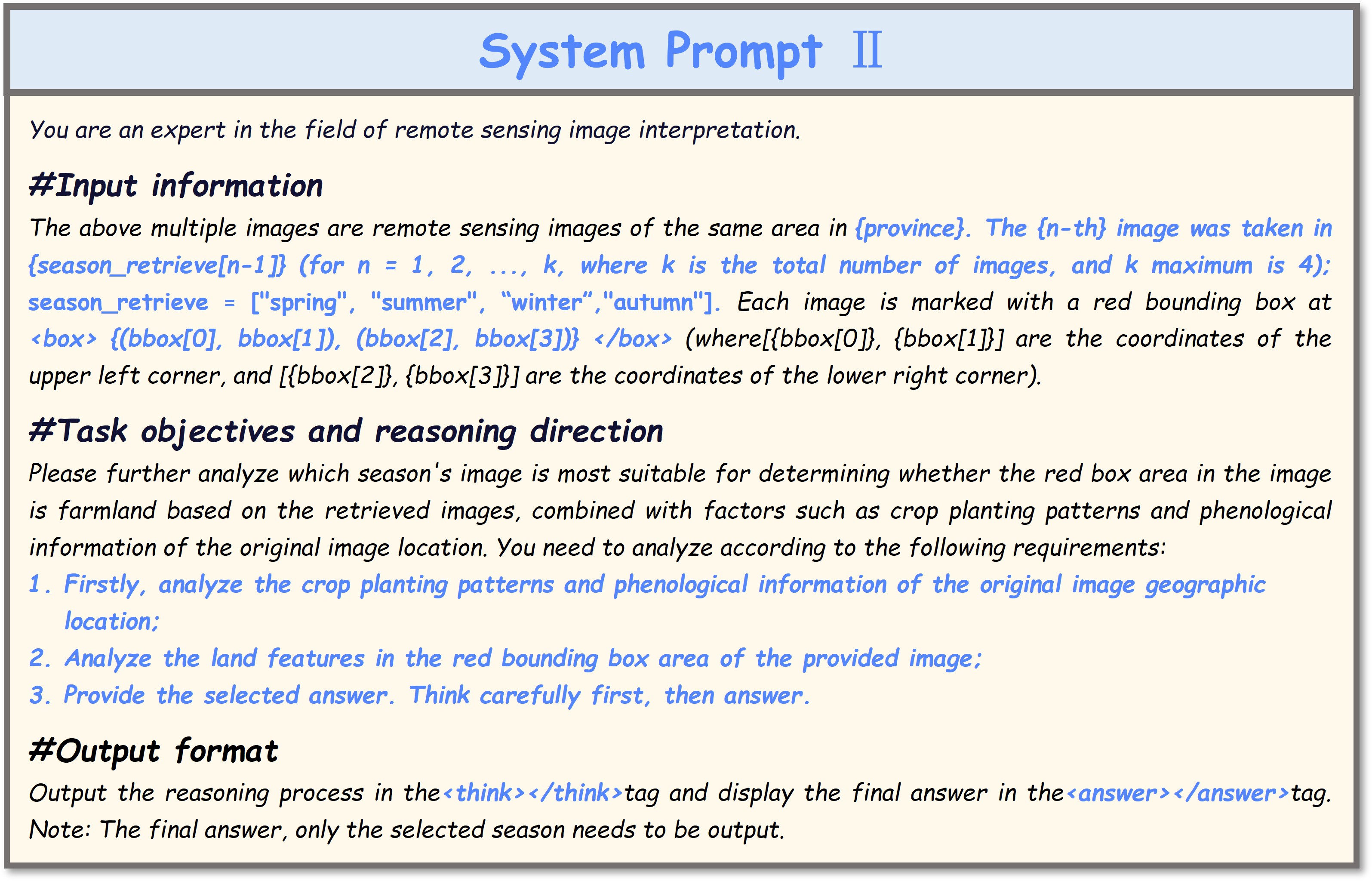}
   \captionsetup{skip=2pt} 
   \caption{System Prompt II ( Template for multi-temporal image retrieval).}
   \label{fig:prompt2(mti)}
\end{figure*}

\begin{figure*}[htbp]
  \centering
  \includegraphics[width=0.8\linewidth]{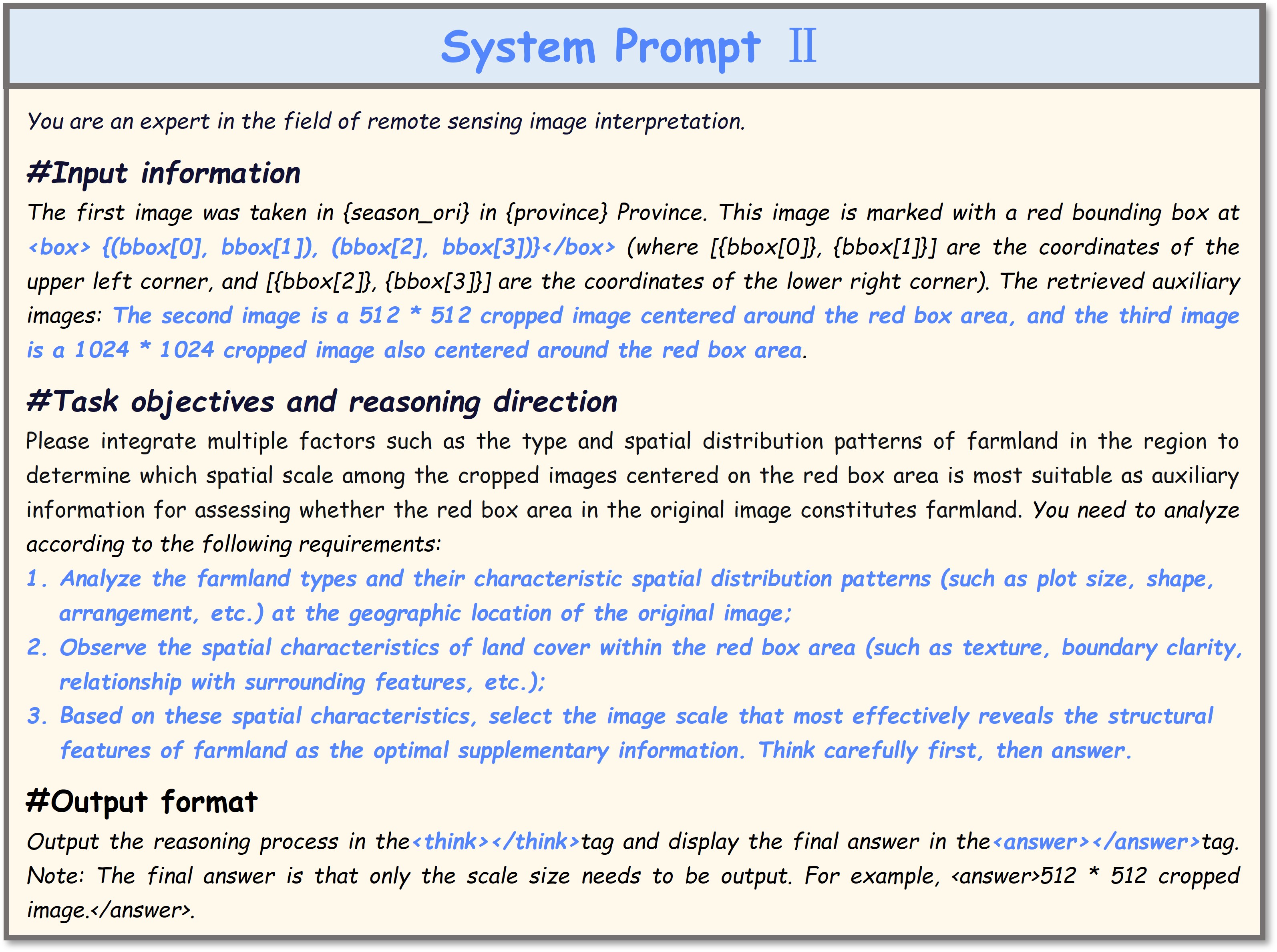}
   \captionsetup{skip=2pt} 
   \caption{System Prompt II ( Template for enlarge image retrieval).}
   \label{fig:prompt2(ei)}
\end{figure*}

In the FarmMind framework, to guide the multimodal large language model (MLLM) in efficiently and accurately identifying farmland, we have designed a hierarchical system prompt. This prompt system divides the overall reasoning process into three levels (System Prompt I, II, and III) based on task complexity and the depth of required information, collaboratively guiding the model to dynamically address three key issues: when to initiate a query, what information to query, and how to effectively utilize the query results:

\begin{itemize}
    \item System Prompt I is designed to guide the MLLM to assess the reasons why the target area in the input image is difficult to accurately judge, and accordingly plan subsequent data retrieval directions. As shown in Figure~\ref{fig:prompt1}, this prompt guides the model to determine whether the difficulty stems from a lack of phenological features or insufficient spatial context, and recommends obtaining multi-temporal images or enlarge images, thereby providing necessary data support for subsequent analysis.
\end{itemize}
\begin{itemize}
    \item System Prompt II is designed to guide the MLLM to evaluate and filter the retrieved auxiliary images, including multi-temporal remote sensing images (see Figure~\ref{fig:prompt2(mti)}) or enlarge images (see Figure~\ref{fig:prompt2(ei)}). This prompt guides the MLLM to formulate discrimination strategies for different types of auxiliary images based on regional crop planting patterns and spatial texture features, selecting from the candidate data the information sources that most effectively reveal the essential attributes of the farmland. This ensures that the final decision is based on the most discriminative and semantically relevant data.
\end{itemize}

\begin{figure*}[htbp]
  \centering
  \includegraphics[width=0.8\linewidth]{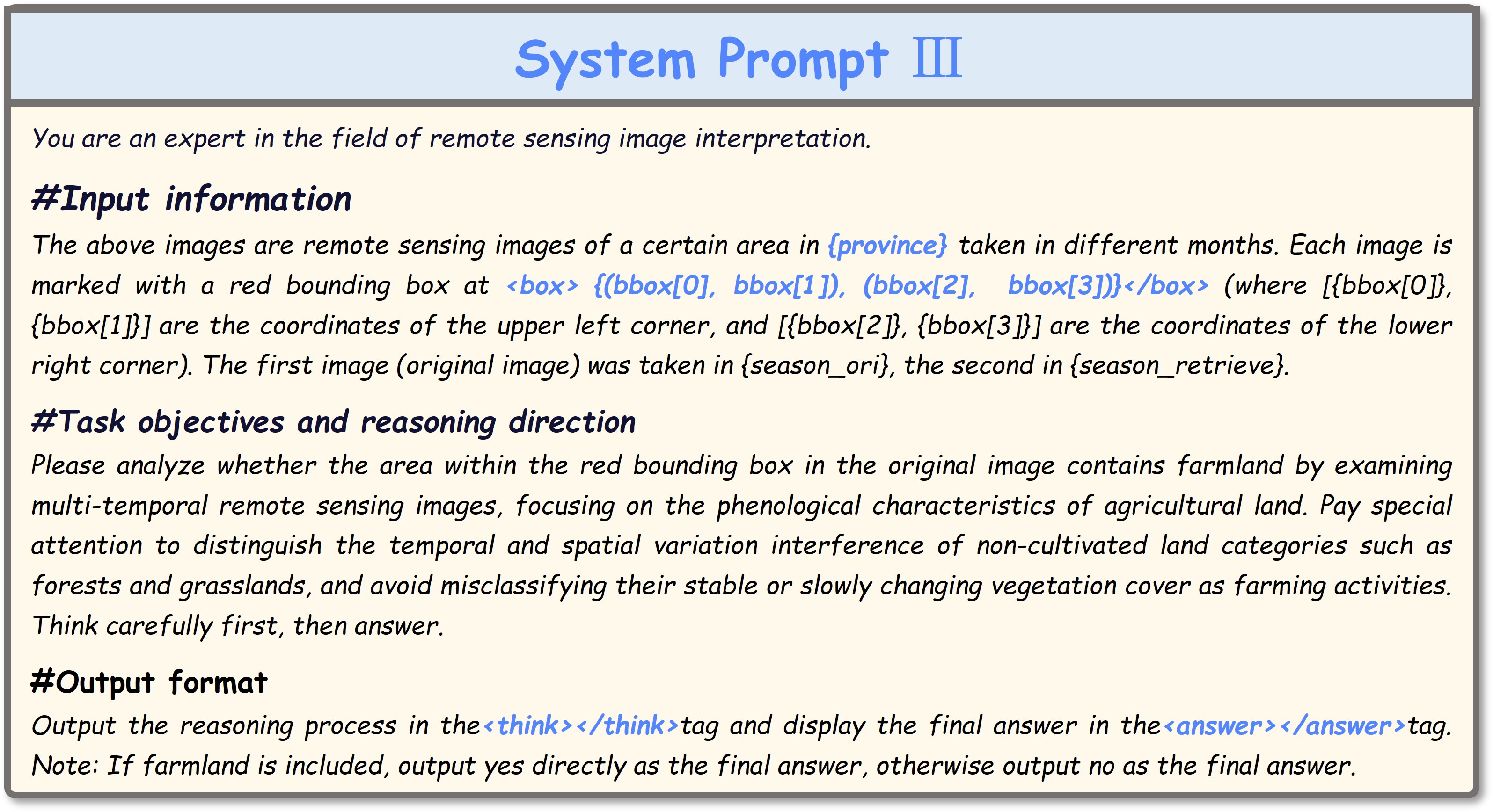}
   \captionsetup{skip=2pt} 
   \caption{System Prompt III ( Template for multi-temporal image retrieval).}
   \label{fig:prompt3(mti)}
\end{figure*}

\begin{figure*}[htbp]
  \centering
  \includegraphics[width=0.8\linewidth]{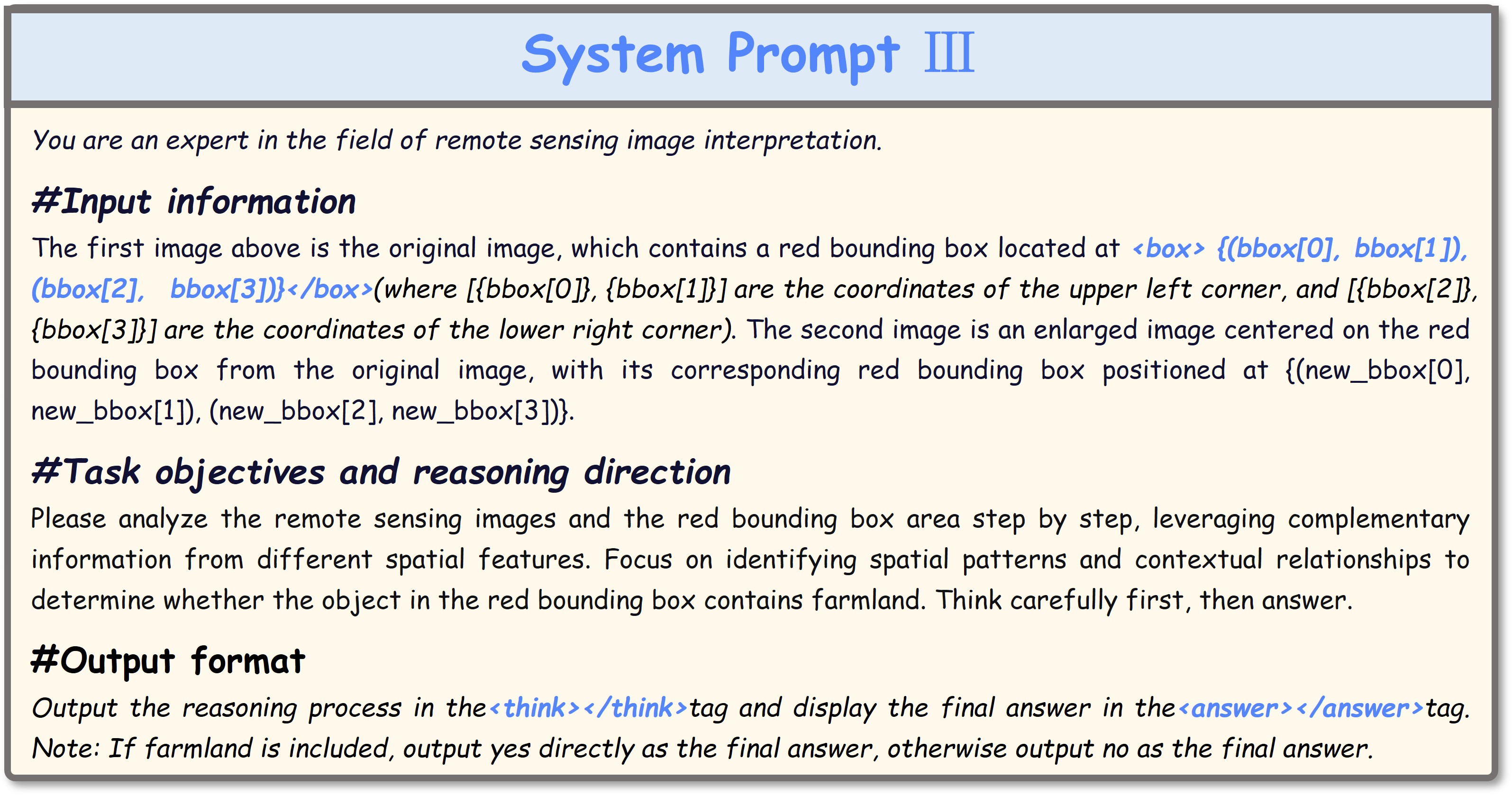}
   \captionsetup{skip=2pt} 
   \caption{System Prompt III ( Template for enlarge image retrieval).}
   \label{fig:prompt3(ei)}
\end{figure*}

\begin{itemize}
    \item System Prompt III is designed to guide the MLLM to better utilize auxiliary images for final collaborative reasoning and perform precise binary judgments. Similarly, this prompt addresses different retrieval types, including multi-temporal remote sensing images (see Figure~\ref{fig:prompt3(mti)}) or enlarged images (see Figure~\ref{fig:prompt3(ei)}), guiding the model to conduct reasoning analysis from different perspectives such as crop phenological characteristics and spatial patterns, to distinguish between farmland and non-farmland.
\end{itemize}

\section{Farmland remote sensing image database details}
\label{sec:C}
\begin{table*}[htbp]
\centering
\renewcommand{\arraystretch}{1.4}
\captionsetup{skip=2pt} 
\caption{Farmland remote sensing image (FRSI) database details.}
\label{tab:SM-tab1}
\begin{tabularx}{\textwidth}{@{}
    >{\hsize=0.5\hsize\centering\arraybackslash}X  
    >{\hsize=2.6\hsize\centering\arraybackslash}X  
    >{\hsize=0.8\hsize\centering\arraybackslash}X  
    >{\hsize=0.5\hsize\centering\arraybackslash}X  
    >{\hsize=0.6\hsize\centering\arraybackslash}X  
    @{}}
\toprule
\textbf{Country} & \textbf{Region} & \textbf{Season} & \textbf{Area(km$^2$)} & \textbf{Resolution} \\
\midrule
China & \makecell[c]{Anhui, Yunnan, Henan, Hebei, Heilongjiang, Sichuan,\\Jilin, Ningxia, Shanxi, Shandong, Guangdong, Hunan} & \makecell[c]{Spring, Summer\\Autumn, Winter} & 4043.7 & 0.5m-2m \\
\cmidrule(lr){2-5}
\makecell[c]{the US} & California & \makecell[c]{Spring, Summer\\Autumn, Winter} & 202.6 & 0.5m \\
\cmidrule(lr){2-5}
Cambodia & Kampot & \makecell[c]{Spring, Summer,\\Autumn, Winter} & 15.7 & 0.5m \\
\cmidrule(lr){2-5}
Germany & North Rhine-Westphalia & \makecell[c]{Spring, Summer\\Autumn, Winter} & 118.8 & 0.5m \\
\bottomrule
\end{tabularx}
\end{table*}

This paper constructs a large-scale, multi-regional, high-resolution farmland remote sensing image (FRSI) database, to support the dynamic segmentation task for farmland remote sensing images. As shown in Table~\ref{tab:SM-tab1}, the database integrates historical remote sensing data from representative agricultural countries and regions, including China, the United States, Cambodia, and Germany. It encompasses various climate zones, crop types, and farming practices, offering substantial geographical and seasonal diversity. In China, the data spans 12 major agricultural provinces, including Anhui, Yunnan, Henan, Hebei, Heilongjiang, Sichuan, Jilin, Ningxia, Shanxi, Shandong, Guangdong, and Hunan, with a total coverage area of 4043.7 km². The spatial resolution ranges from 0.5m to 2m, with data sourced from FarmSeg-VL. In the United States, the focus is on California, covering an area of 202.6 km² at a resolution of 0.5m. In Cambodia, the Kampot region is selected, with a coverage area of 15.7 km² and a resolution of 0.5m. In Germany, the North Rhine-Westphalia region is covered, with an area of 118.8 km² and a resolution of 0.5m. Data from all regions cover all four seasons—spring, summer, autumn, and winter—fully capturing the dynamic changes in crop growth cycles.

\section{Additional experiments and discussions}
\label{sec:D}


\begin{figure*}[htbp]
  \centering
  \includegraphics[width=1\linewidth]{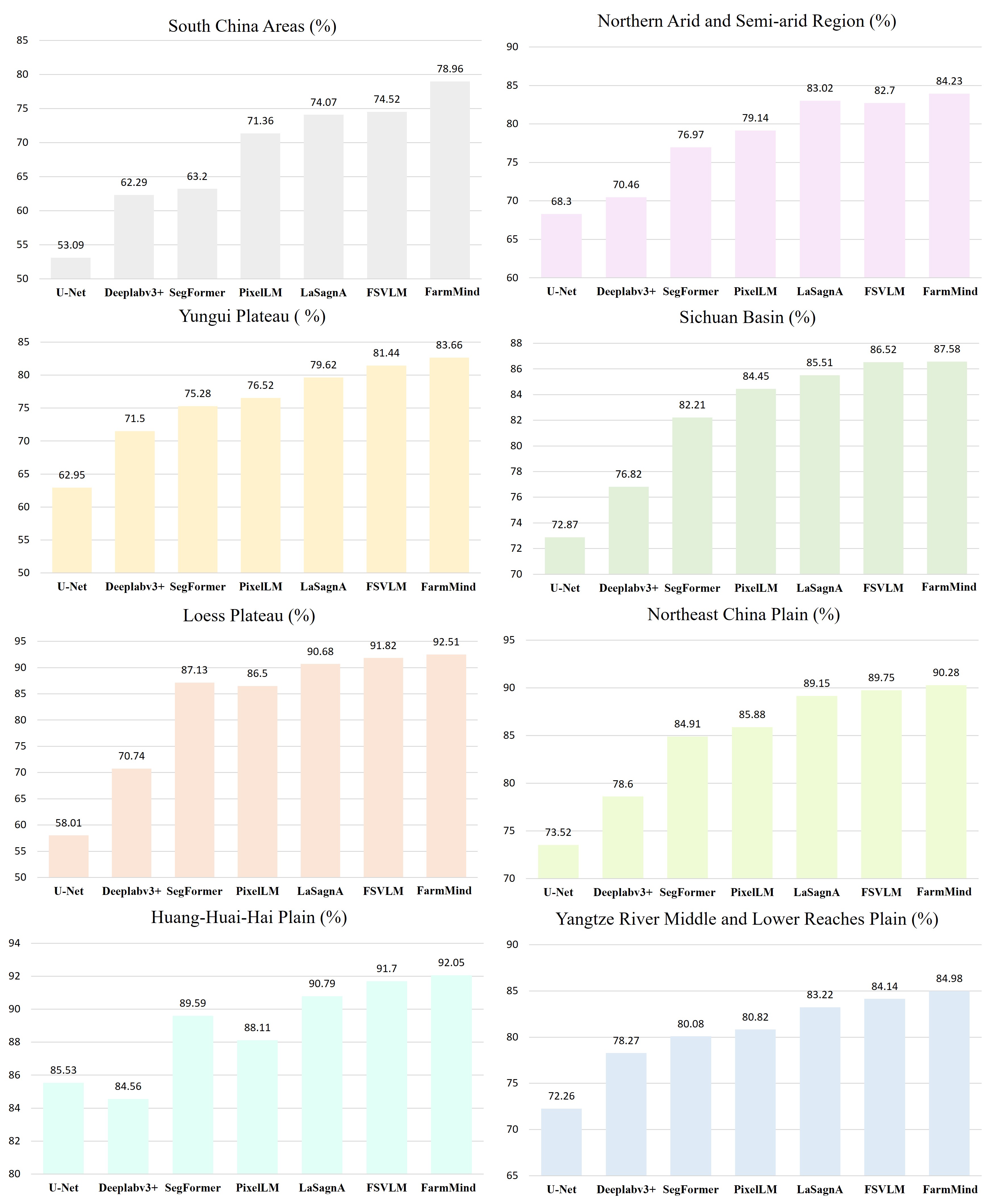}
   \captionsetup{skip=2pt} 
   \caption{The segmentation results of various comparative methods in different agricultural regions (Evaluation metric is mIoU).}
   \label{fig:SM-fig3}
\end{figure*}
\textbf{(1) Segmentation performance of farmMind across different regions in China} 

As shown in Figure~\ref{fig:SM-fig3}, the segmentation results of various comparative methods in different agricultural regions are presented. These regions include the Northeast China Plain (Heilongjiang, Jilin), Huang-Huai-Hai Plain (Hebei, Henan, Shandong), Northern Arid and Semi-Arid Region (Ningxia), Loess Plateau (Shanxi), Yangtze River Middle and Lower Reaches Plain (Anhui, Jiangshu, Hunan), South China Areas(Guangdong), Sichuan Basin(Sichuan), and Yungui Plateau(Yunnan). Specifically, we have chosen mIoU as the evaluation metric here.

The experimental results show that label-driven segmentation methods perform relatively poorly across various regions, particularly in the highly complex and heterogeneous agricultural environments of South China Areas and the Loess Plateau. In contrast, language-driven segmentation methods (such as PixelLM, LaSagnA, and FSVLM) significantly improve segmentation performance in complex farmland layouts, owing to the language reasoning ability that better models the intricate spatial relationships between farmland and surrounding features. The proposed FarmMind method achieves the best performance across all test regions, with particularly notable advantages in regions such as South China Areas, where agricultural structures are highly complex. These areas typically involve dynamic farming patterns, such as crop rotation and mixed planting, which are challenging for traditional static segmentation methods. FarmMind actively queries auxiliary information, such as multi-temporal imagery and enlarge images, to effectively alleviate semantic ambiguities, thereby achieving more refined and robust segmentation. Its consistently excellent performance across diverse regions fully demonstrates the method's outstanding generalization ability and its potential for adaptation to dynamic agricultural environments.

\noindent\textbf{(2) The necessity of reasoning-query}

The reasoning-query mechanism proposed in this paper dynamically analyzes the sources of uncertainty in the model's current predictions and actively determines the types of supplementary information required. For instance, if the model's classification of a region is ambiguous due to a lack of temporal growth state comparisons, historical temporal images of the location are prioritized for retrieval; if the ambiguity arises from insufficient spatial details, high-resolution magnified images of the region are retrieved instead. This "on-demand retrieval" strategy not only improves information utilization efficiency but also enhances the model's adaptability to complex agricultural scenarios.

As shown in the ablation experiment in \cref{table:performance_metrics}, using fixed strategies (such as only retrieving temporal images or only retrieving magnified images) leads to slight performance improvements (mIoU increases from 83.39\% to 83.68–83.81\%), but the gains are limited and lack general applicability. In contrast, the full version of the FarmMind method, which dynamically guides queries through attribution reasoning, achieves significant and consistent improvements across mAcc, mIoU, F1, and Recall metrics (with mIoU reaching 84.43\% and Recall reaching 93.04\%). This clearly demonstrates that the effective introduction of external information must be based on an understanding of the model's internal decision-making process.

\begin{table}[ht]
\centering
\small
\vspace{-9pt}
\caption{Ablation study of reasoning-query on FarmSeg-VL}
\vspace{-9pt}
\resizebox{0.47\textwidth}{!}{
\begin{tabular}{ccccc}
\hline
\textbf{NO.} & \textbf{mAcc (\%)} & \textbf{mIoU (\%)} & \textbf{F1 (\%)} & \textbf{Recall (\%)} \\ \hline
No reasoning query & 90.02 & 83.39 & 90.12 & 90.51 \\ 
Query only temporal images & 91.27 & 83.81 & 90.89 & 92.35 \\ 
Query only enlarge images & 90.58 & 83.68 & 90.72 & 91.51 \\ 
Complete FarmMind & \textbf{92.85} & \textbf{84.43} &\textbf{ 91.25} & \textbf{93.04} \\ \hline
\end{tabular}
\label{table:performance_metrics}
}
\end{table}

\noindent\textbf{(3) Computational cost}

The FarmMind proposed in this paper significantly enhances the semantic understanding and generalization performance of segmentation by incorporating a MLLM for semantic reasoning and interactive correction. However, it inevitably introduces additional computational overhead, primarily due to the remote invocation and inference process of the MLLM. To quantify this impact, we conducted a computational cost experiment: on the FarmSeg-VL dataset, using Qwen-VL-Max as the MLLM backend and remote invocation through Alibaba Cloud API (without occupying local GPU resources), we measured the average per-sample processing time for each module. As shown in Table \ref{tab:my-table}, the majority of the computational time is spent on the MLLM inference. However, it is important to emphasize that the core goal of this paper is to propose a new segmentation framework, not to bind it to a specific model. Therefore, although the above experiment uses the high-performance but computationally intensive Qwen-VL-Max as an example, in practical deployment, users can fully replace it with a lightweight MLLM. The computational overhead is not an inherent flaw of the framework, but a controllable trade-off that can be adjusted according to needs.
\begin{table}[ht]
\centering
\small
\setlength{\tabcolsep}{2pt}
\vspace{-9pt}
\caption{Computational cost}
\vspace{-10pt}
\resizebox{0.47\textwidth}{!}{
\begin{tabular}{@{}cccccc@{}}
\toprule
\multirow{2}{*}{Basic perception (s)} &
\multicolumn{2}{c}{Reasoning query (s)} &
\multicolumn{2}{c}{Collaborative segmentation (s)} \\
\cmidrule(lr){2-3} \cmidrule(l){4-5}
& MLLM reasoning & Query & MLLM reasoning & Dynamic correction \\
\midrule
1.02 & 8.64 & 0.23 & 4.21 & 0.52 \\
\bottomrule
\end{tabular}
\label{tab:my-table}
}
\end{table}

\end{document}